%% file: main.tex
\documentclass[letterpaper, 10 pt, conference]{ieeeconf}  % Comment this line out if you need a4paper

\IEEEoverridecommandlockouts                              % This command is only needed if 
\usepackage[sorting=none, style=numeric-comp, maxbibnames=99]{biblatex}
\input{packages}

\input{macros}

\title{\LARGE \bf
Discovering Synergies for Robot Manipulation\\with Multi-Task Reinforcement Learning
}

\author{Zhanpeng He and Matei Ciocarlie 
\thanks{Zhanpeng He is with the Department of Computer Science, Columbia University, New York, USA.
        {\tt\small zhanpeng@cs.columbia.edu}}%
\thanks{Matei Ciocarlie is with the Department of Mechanical Engineering, Columbia University, New York, USA.
        {\tt\small matei.ciocarlie@columbia.edu}}%
\thanks{This work was supported in part by NSF grant IIS-1551631 and ONR grant N00014-21-1-4010. }
}

\begin{document}

\maketitle
\thispagestyle{empty}
\pagestyle{empty}

%%%%%%%%%%%%%%%%%%%%%%%%%%%%%%%%%%%%%%%%%%%%%%%%%%%%%%%%%%%%%%%%%%%%%%%%%%%%%%%%
\begin{abstract}
Controlling robotic manipulators with high-dimensional action spaces for dexterous tasks is a challenging problem. Inspired by human manipulation, researchers have studied generating and using postural synergies for robot hands to accomplish manipulation tasks, leveraging the lower dimensional nature of synergistic action spaces. However, many of these works require pre-collected data from an existing controller in order to derive such a subspace by means of dimensionality reduction. In this paper, we present a framework that simultaneously discovers both a synergy space and a multi-task policy that operates on this low-dimensional action space to accomplish diverse manipulation tasks. We demonstrate that our end-to-end method is able to perform multiple tasks using few synergies, and outperforms sequential methods that apply dimensionality reduction to independently collected data. We also show that deriving synergies using multiple tasks can lead to a subspace that enables robots to efficiently learn new manipulation tasks and interactions with new objects.
\end{abstract}

%%%%%%%%%%%%%%%%%%%%%%%%%%%%%%%%%%%%%%%%%%%%%%%%%%%%%%%%%%%%%%%%%%%%%%%%%%%%%%%%
\input{text/intro}
\input{text/related_work}
\input{text/method}

\input{text/experiments}

\input{text/results}

\input{text/conclusion}

% \section*{ACKNOWLEDGMENT}

\printbibliography

\addtolength{\textheight}{-12cm}   % This command serves to balance the column lengths
\end{document}

%% file: packages.tex
\usepackage{color}
\usepackage{epsfig}
\usepackage{graphicx}
\usepackage{algorithm,algorithmic}
\usepackage{adjustbox}
\usepackage{array}
\usepackage{booktabs}
\usepackage{colortbl}
\usepackage{float,wrapfig}
\usepackage{framed}
\usepackage{hhline}
\usepackage{multirow}
\usepackage{subcaption} % issues a warning with CVPR/ICCV format
\usepackage[font=small]{caption}
\usepackage[percent]{overpic}
\usepackage{amsmath,amsfonts,amssymb}
\usepackage{amsthm} 
\usepackage{bm}
\usepackage{nicefrac}
\usepackage{microtype}
\usepackage{contour}
\usepackage{courier}
\usepackage{changepage}
\usepackage{extramarks}
\usepackage{fancyhdr}
\usepackage{lastpage}
\usepackage{setspace}
\usepackage{soul}
\usepackage{xspace}
\usepackage{cuted}
\usepackage{fancybox}
\usepackage{afterpage}
\usepackage[breaklinks=true,colorlinks,backref=True]{hyperref}
\hypersetup{colorlinks,linkcolor={black},citecolor={green},urlcolor={magenta}}
\usepackage{url}
\usepackage{quoting}
\usepackage{epigraph}

\usepackage{enumerate}
\usepackage{paralist,tabularx}
\usepackage{comment}
\usepackage{pdfpages}
\usepackage{caption}
\usepackage{subcaption}

\usepackage{pifont}% http://ctan.org/pkg/pifont

%% file: macros.tex
\usepackage{enumitem}

\makeatletter
\DeclareRobustCommand\onedot{\futurelet\@let@token\@onedot}
\def\@onedot{\ifx\@let@token.\else.\null\fi\xspace}

\def\etal{et al\onedot}

\makeatother

\definecolor{MyDarkBlue}{rgb}{0,0.08,1}
\definecolor{MyDarkGreen}{rgb}{0.02,0.6,0.02}
\definecolor{MyDarkRed}{rgb}{0.8,0.02,0.02}
\definecolor{MyDarkOrange}{rgb}{0.40,0.2,0.02}
\definecolor{MyPurple}{RGB}{111,0,255}
\definecolor{MyRed}{rgb}{1.0,0.0,0.0}
\definecolor{MyGold}{rgb}{0.75,0.6,0.12}
\definecolor{MyDarkgray}{rgb}{0.66, 0.66, 0.66}

\newcommand{\bE}{\mathbb{E}}

\newcommand{\cmark}{\ding{51}}%
\newcommand{\xmark}{\ding{55}}%

\newcommand{\head}[2]{\multicolumn{1}{>{\centering\arraybackslash}p{#1}}{#2}}

%% file: text/intro.tex
\section{INTRODUCTION}

Recent advances show promise towards building competent robots hands that are able to achieve complex manipulation tasks.
Many of these robots are designed to be versatile for general manipulation tasks and have numerous degrees of freedom.
For example, the Shadow Hand can accomplish various in-hand manipulation tasks: finger pivoting, sliding, and gaiting a cube~\cite{openai2018dex}.
However, using robots with such high-dimensional control spaces remains a challenge in both model-based control and model-free methods. 
For model-based methods, such as model-predictive control, it is difficult to derive an accurate model due to the manipulators' high degrees of freedom. Even when an accurate model exists, model-based methods are computationally expensive when using high-dimensional state spaces.
In the case of model-free methods, even when it is indeed possible to learn robust policies for manipulation tasks, the training process still requires large amounts of robot experience to accommodate the large action space of highly dexterous hands. Finally, highly actuated hands are difficult and expensive to manufacture, and can be fragile in use. From all of these perspectives, using robot hands with a very high dimensional control space remains a challenge. 

% \begin{figure}[h]
%      \centering
%      \begin{subfigure}{0.68\linewidth}
%          \includegraphics[width=\linewidth]{figures/synergies_hand.pdf}
%      \end{subfigure}
%      \begin{subfigure}{0.3\linewidth}
%          \centering
%          \includegraphics[width=\linewidth]{figures/discosyn-hand-interact-2tasks-vertical.pdf}
%     \end{subfigure}
%     \caption{Visualization of learned synergies (Left) from task set \#1 with 3 synergies. Each color in this plot represents a dimension of the lower dimensional action space. We also visualize some moments of our agent utilizing synergies to solve one of the valve turning task (Right). Our agent learns to gait and turning the valve counter-clockwise. The synergies learned help the agent use different handles of the valve and transition to the next gaiting cycle.}
%     \label{fig:handposes}
% \end{figure}
\begin{figure}[h]
     \centering
     \includegraphics[width=\linewidth]{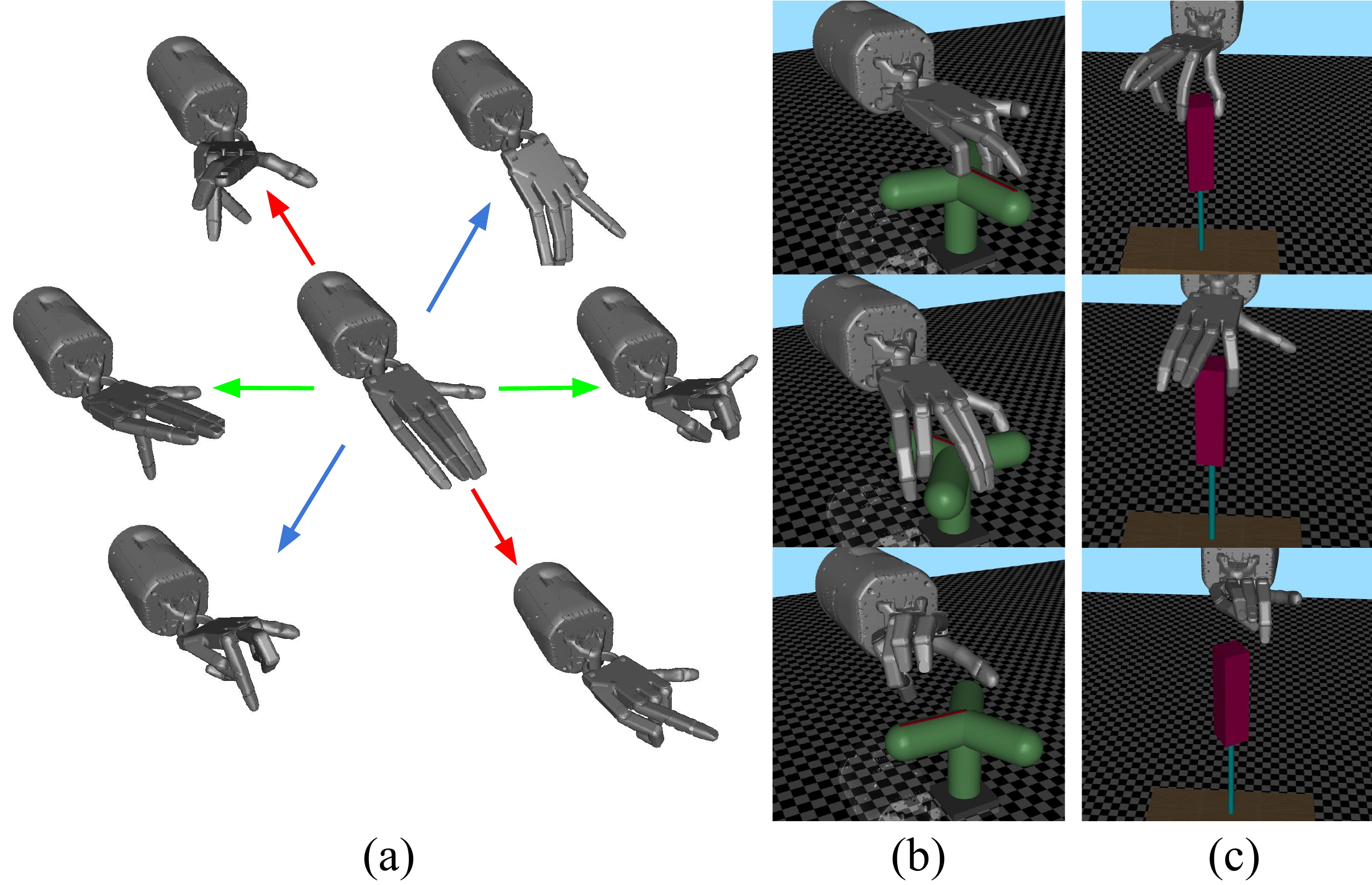}
    \caption{Example of multi-task synergy discovery. The DiscoSyn algorithm can discover manipulation synergies such as the 3-dimensional linear synergy set shown in (a), while simultaneously deriving control policies that use these synergies for manipulation tasks such as the valve turning task shown in (b). Learned synergies also facilitate subsequent learning of new manipulation tasks, such as the top-down screwdriving task shown in (c). }
    \label{fig:handposes}
    \vspace{-7mm}
\end{figure}

In contrast, humans can achieve many manipulation tasks in a synergistic way. 
They can achieve efficient object interactions by selecting hand postures from a small configuration space \cite{Napier1956ThePM}\cite{iberall1997humanprehension}\cite{marco1998toolusepose}, even though human hands can reach a large number of postures.
Inspired by human manipulation, roboticists have  also explored the concept of motor synergies in the context of robotic manipulation, looking to find a low-dimensional subspace of postures that allow for efficient planning for grasping tasks.
Such a subspace has generally been extracted by performing dimensionality reduction on ample amounts of task solution data collected via simulations or human demonstrations.
% Specifically, they collect solutions to some robotic tasks, which are high dimensional, via simulations or human demonstrations. Then, synergies are generated by applying Principal Component Analysis (PCA) on these high dimensional data. 

However, collecting such data for dimensionality reduction can be slow and expensive since this process requires an expert with domain knowledge of the robotic task to generate solutions. 
Even when such data exist, the extracted synergies can be biased if the solutions that they are learned on lack variety. Furthermore, although human hands inspire many designs of robot manipulators, these have different kinematic structures, and robots do not necessarily accomplish tasks in the same ways as humans. Therefore, synergies generated from human collected data might not be helpful for robot manipulation, and discovering synergies directly using robotic hands remains an essential problem. 

% \begin{figure}
%     \centering
%     \includegraphics[width=\linewidth]{figures/synergies_hand.pdf}
%     \caption{}
%     \label{fig:handposes}
%     \vspace{-6mm}
% \end{figure}

While many works on synergies focus on grasping, finding such a subspace can be helpful in many others manipulation tasks, and such a subspace could potentially be shared across multiple tasks.
For example, using a screw driver from a top-down pose and turning a cylindrical dial can be achieved using similar hand postures to grasp the target object.
Learning synergies from a number of diverse tasks exhibiting complex dynamics could lead to a subspace that in turn contains diverse hand postures. Such a subspace can potentially provide efficient exploration while interacting with new objects and hence allow for learning a control policy faster for an unseen task.

% Having a library of synergies that are shared across multiple tasks allows for efficient task learning. 
% A recent work \cite{han2021synergy} on reinforcement learning shows the emergence of synergy in learning arm manipulation tasks in a single-task setting. 
% However, discovering and extracting synergies that are useful for diverse tasks remains a question. 
% Such synergies can provide efficient exploration when we try to learn a policy for an unseen task that can be achieved with similar postures. 

Finally, although linear synergies are well studied in the field of grasping, they can have limited applicability since linear models have limited capacity. We would like to discover linear synergies when feasible, but also allow for the flexibility of learning non-linear synergies which are more capable of expressing diverse hand postures.

% Furthermore, in the field of manipulation, most works only explore linear synergies. Although linear synergies can be interpreted by humans and are easy to analyze, linear models can store limited information compared to more complex models, for example, deep neural networks. Learning nonlinear synergies leads to a subspace that contains diverse hand postures that can be potentially useful for more manipulation tasks.

In this work, we present DiscoSyn, a framework that simultaneously discovers a synergy space and control policies that leverage this space for manipulation. 
Instead of extracting hand synergies from pre-collected control data, our method builds on multi-task reinforcement learning, which allows us to learn policies for multiple tasks that can share a synergy space.
The results of our framework include both a policy that selects task-specific sequences of low-dimensional actions, and a synergy model shared across all tasks which projects these synergistic actions back to the original high-dimensional action space.
%The hope is that by learning synergies from enough tasks, this subspace contains rich postural synergies that can be reused and provide an agent a small state space that allows for efficient exploration while encountering new tasks.
We summarize our contributions as follows:
\begin{compactitem}
    \item To the best of our knowledge, we are the first to propose an algorithm that can discover synergies shared across multiple manipulation tasks and simultaneously learn policies that select actions from the synergy space.
    \item We demonstrate that our framework can learn both linear synergies, which have the advantage of being easily interpretable, and non-linear synergies, which provide a larger space that can be more suitable for learning diverse manipulation tasks.
    \item We show evidence that the learned synergies allow an agent to explore new environments efficiently and can be used to learn previously unseen tasks. 
\end{compactitem}

%% file: text/related_work.tex
\begin{figure*}
    \centering
    \includegraphics[width=0.75\linewidth]{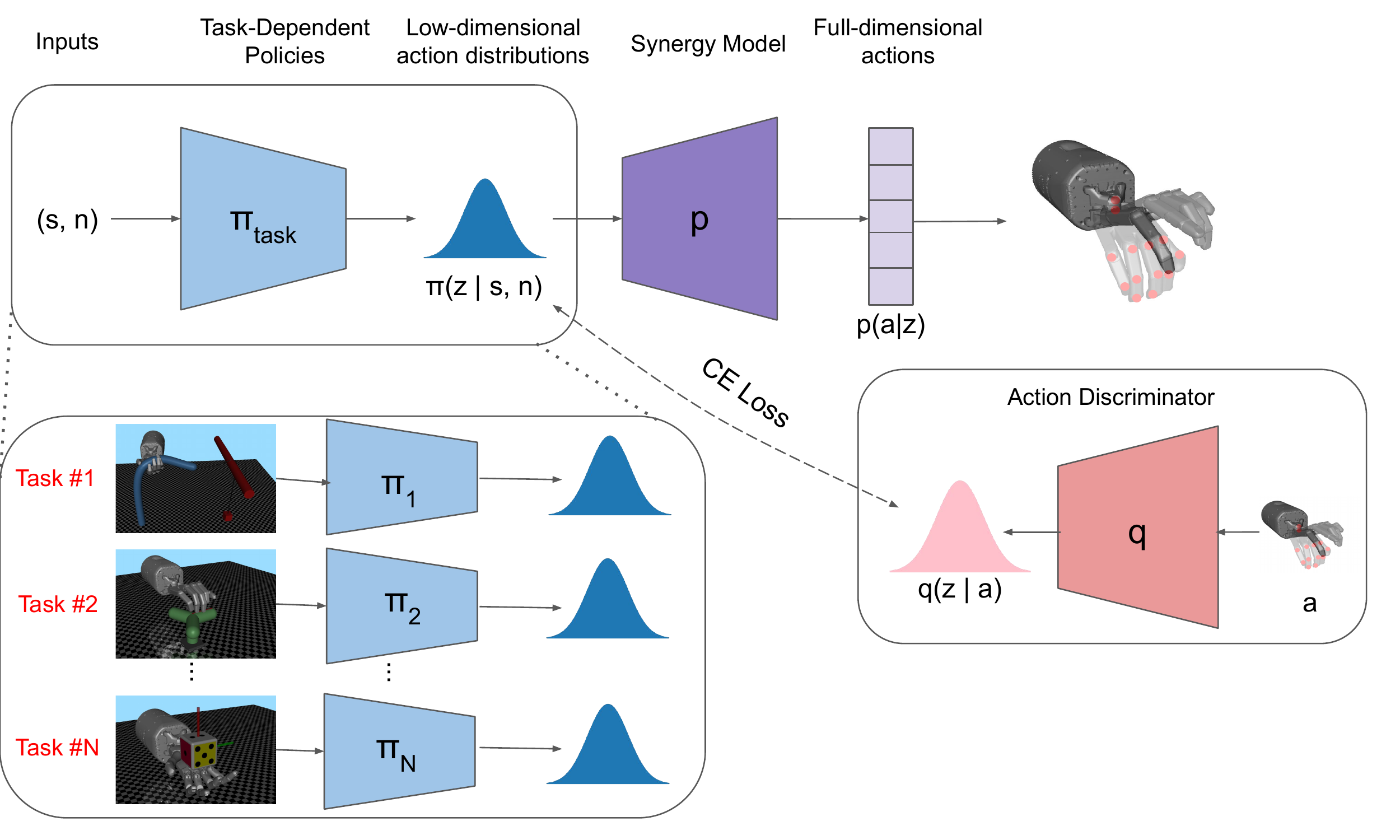}
    \caption{Overview of DiscoSyn. Observations from each task are fed to the task-dependent policy head $\pi_{task}$ to derive low dimensional actions $z$. Then, the synergy model $p$ takes the $z$ as inputs and infer the full dimensional actions $a$ that can be directly applied to the environment. }
    \label{fig:disco_syn}
    \vspace{-6mm}
\end{figure*}

\section{Related Work}

Inspired by human-like manipulation, robotics researchers have attempted to leverage synergistic manipulation for a wide range of applications.
Several works \cite{marco1998toolusepose}\cite{mason2002reachgraspsyn}\cite{todorov2004handsynergies}\cite{Liu2016AnalysisOH}\cite{Patel2017HandGS}\cite{nester2019hansynergydata}\cite{ciocarlie2009posture}\cite{suarez2015dualarm} applied dimensionality reduction on hand postures to study this behavior. 
These works follow the same sequential pipeline:  their approaches first gather high-dimensional actions and extract synergies using by applying dimensionality reduction techniques, such as Principal Component Analysis (PCA), and show that a few principal components (PC's) can explain most of the variance. 
However, these only study linear synergies and most of them only work on grasping tasks. 
Our work, in contrast, focuses on control synergies of robotic hands that are useful for multiple dexterous manipulation tasks and do not place constraints of the form of synergies. Another key feature of our framework is that it does not rely on pre-collected data. Our agent finds solutions for multiple tasks when discovering synergies.

To find solutions for robotic tasks automatically, our framework utilizes reinforcement learning, which has shown promising results in learning different manipulation tasks \cite{kroemer2019manipulation}\cite{Daniel2016reps}\cite{peter2008rlmotorskills}\cite{theodorou2010reinforcementklo}. 
Nagabandi \etal \cite{nagabandi2019pddm} propose a model-based method that learns a dynamics model of the environment and then multiple tasks such as a Baoding ball, weight pulling, and valve turning. 
OpenAI \etal \cite{openai2018dex} demonstrate the effectiveness of model-free methods, such as Proximal Policy Optimization (PPO)\cite{schulman2017ppo}, on learning dexterous in-hand manipulation. However, all of these works learn policies on the full-dimensional action space. Our work leverages these advances of RL algorithms, but also allows an agent to discover a low-dimensional action space that can be used to learn new tasks efficiently.

To discover synergies across different tasks, we place our agents in multiple environments.
Researchers have extensively studied learning multiple tasks simultaneously using RL and find that multi-task RL can leverage shared information across different tasks \cite{Tanaka2003mtmdps}\cite{Calandriello2014sparsemtrl}\cite{borsa2016mtrl}\cite{DEramo2020Sharing}. 
Hausman \etal \cite{hausman2018learning} proposed a framework that learns continuous task representations and a task-dependent policy. Yu \etal \cite{yu2019meta} investigate the performance of various multi-task RL algorithms; for example, multi-task soft actor-critic, on several multi-task sets (Meta-world).
Sodhani \etal \cite{sodhani2021contextrl} present a method that leverages languages as context to learn representations of a task to facilitate learning of multiple tasks. 
All of these works show promising results on learning multiple tasks considering extracting prior knowledge to learn unseen tasks.
To achieve this, many previous works show the generalization ability of their task representation and try to find an appropriate representation for an unseen task. 
Our work, focused on robot hands, proposes a synergy model that contains hand configurations that allow for efficient exploration of unseen manipulation tasks and objects. 

The closest work to ours is Group Factor Policy Search (GrouPS)\cite{luck2017groups}, which integrates Group Factor Analysis (GFA) with RL. GrouPS designs the policy to be of a particular structure, which can be further interpreted as a linear synergy model, and directly applies policy search using this policy.  While it shows promising results on extracting synergies from a two-arm grasping task, it can only use a linear policy in a single task setting. On the other hand, our framework provides the flexibility of choosing the structure of both the task-dependent policy and the synergy model and we are thus able to apply it to more complex manipulation tasks. While GrouPS focus on learning synergies on a single grasping task, our work emphasizes the importance of extracting synergies from multiple manipulation tasks that require more dexterous motor skills.

%% file: text/method.tex
\section{Approach}
% High dimensional action space
% low dimension action space
% synergy -- decoder
% linear -- decoder is just a matrix
% policy that chooses action in the synergy space
% co optimization
% then formalism

Our method aims to extract control synergies among diverse tasks which provide a compact subspace for agents to derive a policy efficiently. We employ a general definition of a synergy -- a manifold that encodes a subset of high-dimensional control commands. A point $z \in \mathcal{R}^b$ in synergy space represents a low-dimensional action, which can then be projected to a point $a \in \mathcal{R}^d$ in the original high-dimensional action space using a synergy model $p(a | z)$. The parameters $\phi$ of this model effectively determine the synergy space. For example, if we choose to use a deterministic linear synergy model, then $\phi \in \mathcal{R}^{b \times d}$ encapsulates a matrix, and $p$ represents the multiplication of $z$ by $\phi$. However, the general notation introduced here allows for more general, potentially non-linear or non-deterministic synergy models. The first key goal of our approach is thus to learn the parameters $\phi$ that determine the behavior of a synergy model appropriate for a given set of manipulation tasks.

In addition to learning the synergy model $p(a | z)$, the second key goal of our framework is to simultaneously learn task-conditioned policies that find task solutions by generating sequences of low-dimensional actions $z$. Critically, we force our agent to discover shared synergies among different tasks by only using one synergy model to decode these low-dimensional solutions.

In summary, our agent takes observations from a task, and generates an low-dimensional action with a task-conditioned policy. This synergistic action is then projected back into a full-dimensional action using a synergy model shared across all tasks. Finally the full-dimensional action is applied to the environment.

% Our method builds on multi-task RL algorithms, which allows agents to discover solutions for multiple tasks. In general multi-task RL, an agent using high-capacity models can produce arbitrary solutions that do not necessarily have a shared structure. 
% Our framework forces an agent to learn solutions that utilize synergies among different manipulation tasks by splitting the policy into a task-conditioned part that operates on the low dimensional action space and a task-independent component shared across all the tasks. In this section, we first define the problem of multi-task learning, then formally describe the integration of synergy learning and multi-task RL, and finally, describe some implementation details of our method.

\subsection{Multi-task Reinforcement Learning}
\label{sec:mttask-rl}
To realize this framework, we train our agent via multi-task RL in multiple Markov Decision Processes (MDP's). An MDP is identified by a tuple of $(\mathcal{S}, \mathcal{A}, r, p_{tran})$, where $\mathcal{S} \in \mathbb{R}^{m}$ represents the state space, and $\mathcal{A} \in \mathbb{R}^{l}$ denotes the action space, $r$ is the reward function $r(s, a)$ and $p_{tran}$ is the probability of transitioning to the next state $s'$: $p_{tran}(s' | s, a)$. We formally define our problem of multi-task RL as follows. 
We assume that we have access to a set of tasks $\mathcal{N} = [1, 2, ..., N]$, which may vary in any aspects of a standard MDP. 
The goal of multi-task RL is to learn a task-conditioned policy $\pi(a | s, n)$ that maximizes the average expected returns across all the tasks, or $\mathbb{E}_{n \sim \mathcal{N}}[\mathbb{E}_{\pi}[\sum^{\infty}_{t=0} \gamma _{t} r^n(a_t, s_t)]]$, where $\gamma_{t}$ is the discounted factor at time step $t$.

\subsection{Synergy learning with multi-task RL}
\begin{figure*}[h]
    \vspace{2mm}

    \centering
    \includegraphics[width=\linewidth]{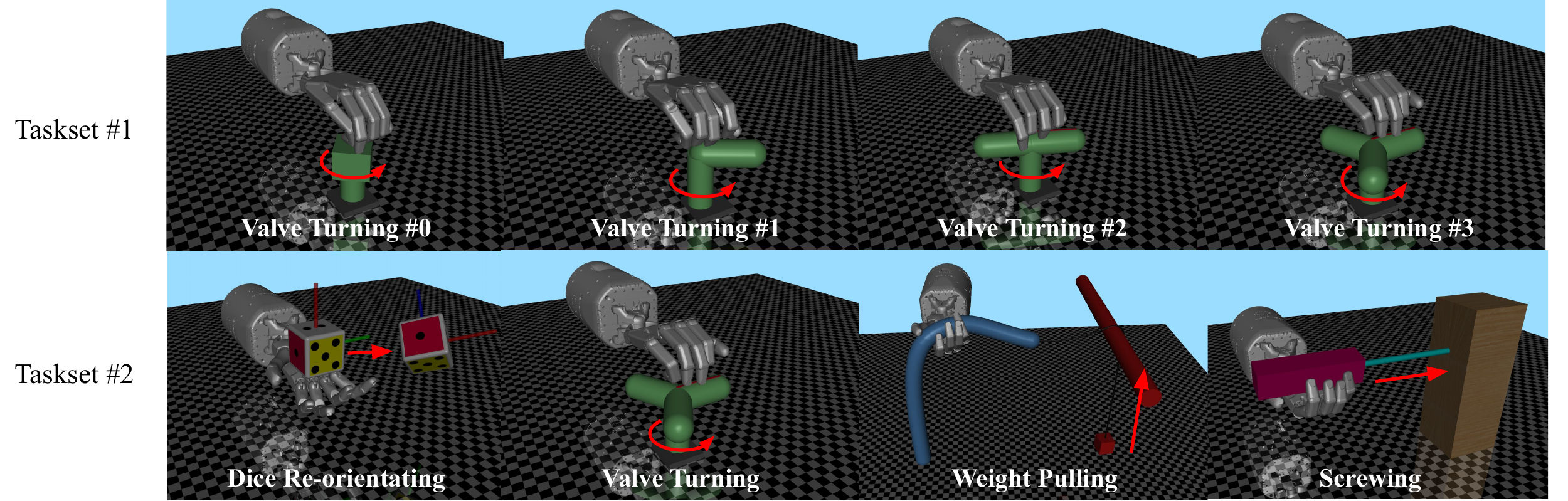}
    \caption{Two task sets are used to test DiscoSyn. Task set \#1: Turning valves with different number of handles. Task set \#2: Four tasks that have different goals (represented as different reward functions) and different dynamics (represented as different transition functions).}
    \label{fig:tasks}
    \vspace{-6mm}
\end{figure*}
\label{sec:discosyn_mttrl}
To integrate synergy learning with multi-task RL, our policy comprises a tasks-conditioned component $\pi_{task}$, whose outputs are low-dimensional actions $z$, and a synergy model $p$ that takes low-dimensional actions and recovers the full dimensional actions $a$: $p(a | z)$. Hence, as shown in Fig. \ref{fig:disco_syn}, an action can be computed from a state and the task identity by $\pi(a| s, n) = \pi_{task}(z| s, n)p(a | z)$.

Our goal is to discover a synergy space that is compact but still contains diverse postures that are efficient for learning different tasks. While learning hand synergies with multiple tasks can potentially lead to such a diverse space, we encourage our policy to learn diverse solutions in the full-dimensional action space of each task by employing maximum-entropy RL.
Specifically, instead of optimizing for the general multitask expected return, we optimize for a maximum-entropy expected return: $
\max_{\pi} \bE_{\pi, n \in \mathcal{N}} \Big[ \sum_{t=0}^\infty \gamma^i \big( r^n(s_t, a_t) + \alpha \mathbb{H}[\pi_\theta(a_t|s_t, n)] \big) \Big]$. 

By itself, the objective function does not provide intuitions about optimizing the synergy space. However, we can derive a lower bound\footnote{A detailed proof is included in \href{https://roamlab.github.io/discosyn/}{https://roamlab.github.io/discosyn/}.} of the entropy term $\mathbb{H}[\pi_\theta(a_t|s_t, n)]$ by applying Jensen's Inequality:
\vspace{-1mm}
\begin{align}
\mathbb{H}[\pi_\theta(a_t|s_t, n)] &= \bE_{\pi}[-\log \pi_\theta(a_t|s_t, n)] \nonumber
\\& \geq \bE_{\pi_\theta(a, z | s, n)} \Big[ \log(\dfrac{q(z| a, s, n)}{\pi(a, z| s, n)}) \Big] \nonumber
\\& = - \bE_{\pi_\theta(a, z | s, n)} \Big[ - \log q(z| a_t) \Big]
\label{eq:lower-bound}
\\& +\mathbb{H} \lbrack \pi_\theta(z_t|s_t, n) \rbrack + \bE_{\pi_\theta(z|s, n)} \Big[ \mathbb{H}\lbrack p_{\phi}(a_t|z_t) \rbrack \Big]
 \nonumber
\end{align}

Equation \ref{eq:lower-bound} gives us an intuition on how to optimize for such a lower action space. 
The first term suggests that the low dimensional actions should be identifiable by the full-dimensional actions. The second term and third term encourage our agent to find diverse solutions for each task. Since the true distribution $q(z|a)$ is not tractable, we approximate it by learning a discriminator $q_{\psi}(z|a)$ from sampled data. In summary, we optimize our task-conditioned policies and the synergy model using an extended reward:
\vspace{-4mm}

\begin{align}
    \label{eq:final-objective}
    \mathcal{L(\theta, \phi)} = \max_\pi \:\mathbb{E}_{\substack{\pi(a, z|s, n) \\ t\in T}} &
        \left[
            \sum_{i=0}^\infty \gamma^i \hat{r}(s_t, a_t, z_t, n)
        \right]
        \vspace{-1mm}
    \intertext{where}
    \vspace{-1mm}
    \nonumber
    \hat{r}(s_t, a_t, z_t, n) = r^n(s_t, a_t) &+ \alpha_1 \mathbb{E}_{n\in \mathcal{N}} \left[\mathbb{H}\left(\pi_\theta(z_t|s_t, n)\right)\right]\\
      &+ \alpha_2 \log q_\psi(z_t | a_t) \nonumber\\
      &+ \alpha_3 \mathbb{H}\left(p_\phi(a_t | z_t)\right) %\nonumber
\end{align}

Here, $\alpha_{1}$, $\alpha_{2}$, $\alpha_{3}$ are constant. Algorithm \ref{alg:discosyn} shows the procedure of co-optimizing a task-dependent policy and a synergy model. We use PPO to optimize $\pi$ and $p$.
\vspace{-1mm}

\begin{algorithm}
\caption{DiscoSyn training}
\label{alg:discosyn}
\begin{algorithmic}[1]
    \WHILE {returns have not converged}
        \STATE Sample a batch of tasks $n \in \mathcal{N}$ \\
        \STATE Sample $H$ trajectories using the current policy $\pi_{task}(z | s, n)p(a|z)$ for each sampled task \\
        \STATE Optimize discriminator $q$ using collected $(z, a)$\\
        \STATE Optimize $\pi_{task}$ and $p$ with Eq. \ref{eq:final-objective} \\
    \ENDWHILE
\end{algorithmic}
\end{algorithm}

\vspace{-4mm}

\subsection{Implementation Details}
\label{sec:implentation_details}

\subsubsection{Task-conditioned policy} One naive implementation of the task-conditioned policy $\pi_{task}$ can be a single model whose inputs combine the state $s$ and task identity $n$. However, we empirically found that using a single model fails to learn all the tasks.
This can be caused by conflicts in gradients since the tasks we include in our task set require an agent to perform diverse motor skills. Hence, we employ a multi-head structure for $\pi_{task}$, including $N$ models while training our agent with $N$ tasks. When our agent encounters task $n$, we pick the $n$th model and feed the observation input to derive a low dimensional action $z$. %This policy can be directly optimized in a multi-task RL framework that is described in Section \ref{sec:mttask-rl}.

\subsubsection{Synergy model} The framework presented so far does not specify the concrete form of the synergy model $p(a|z)$. We have used it in this work to learn two forms of synergies: linear and non-linear.

Linear synergies produce a manifold that can be interpreted analytically and can be useful in robotic hand design since they provide intuitions for underactuation mechanisms. To learn linear synergies, we employ the dimensionality reduction view also used in other works and treat our synergy model as a reversed process of dimensionality reduction. 

We parameterize our linear synergy models with a matrix $\phi \in \mathcal{R}^{b \times d}$. In the case of using a deterministic synergy model, we can derive a full-dimensional action $a$ by multiplying $z$ by $\phi$. In this work, during training, we use stochastic synergy models to encourage exploration. Thus, we treat the product of $z$ and $\phi$ as parameters of the full-dimensional action distribution. Specifically, in our experiments, we use normal distributions and calculate the mean by $a_{mean} = z \cdot \phi$. Then, we model the standard deviation using a vector in the same shape as $a_{mean}$. During testing, we remove the stochasticity of our synergy model by only using $a_{mean}$ as our action output $a$ for stable hand behaviors.
% In PCA-based synergy extraction, each synergy $e_i$ is a $d$-dimensional vector where $d$ is the dimension of the full action space: $e_i = [e_{i, 1}, e_{i, 2} ... e_{i, d}]$. 
% Then, to recover an action in joint space, we choose $b$ synergy and apply the motion on each synergy direction: $a = \sum^{b}_{i=0}z_{i}e_{i}$. 

On the other hand, although non-linear synergies are difficult to interpret, non-linear models have a larger capacity to produce diverse hand postures, potentially increasing versatility. To learn non-linear synergies, we can use any non-linear models for $p(a|z)$. Here, we use multi-layer perceptrons (MLP) to learn non-linear synergies.

\subsection{Learning unseen tasks using learned synergy model}

DiscoSyn presents a method to learn a hand configuration space for robot manipulation tasks. This space is generally of much smaller dimensionality than the original full dimensional action space, and thus provides an opportunity to speed up learning of new tasks, \textit{i.e.} tasks not included in the original set that the synergy model was learned on. To learn a previously unseen task $n'$, where $n' \notin \mathcal{N}$, we take a learned synergy model $p(a|z)$, freeze the parameters of $p$, and directly optimize a task-dependent policy $\pi(z|s, n')$, whose outputs are lower-dimensional actions $z$.

%% file: text/experiments.tex
\section{Experiments and results}

\subsection{Synergies discovery and baselines}

To test if our method can discover synergies for multiple tasks, we apply DiscoSyn on two training task sets.
As shown in Fig \ref{fig:tasks}, each task set contains four manipulation tasks; we will discuss their design considerations below. Although these tasks have different observation spaces, we use a multi-headed structure network for the task-dependent policy $\pi_{task}(z | s, n)$ (see Sec. \ref{sec:implentation_details}).
Hence, the input dimension can be different for each head.
We also test our method using different numbers of dimensions for the $z$ space, which corresponds to the number of synergies we extract. 
We evaluate the policy learning performance, as well as the ability to extract low dimensional representations for the high dimensional actions, along with the explained variance of the data.

\begin{figure}[t]
    \includegraphics[width=0.95\linewidth]{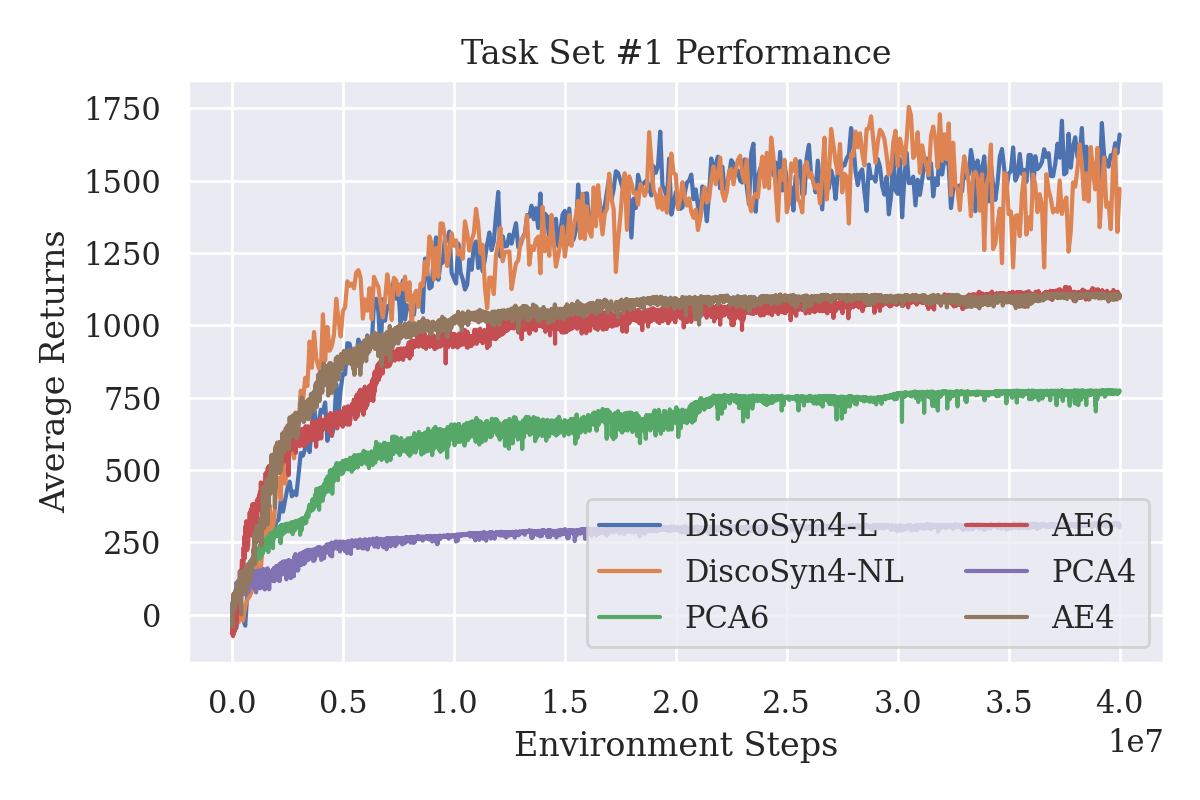}
    \vspace{-2mm}
    \caption{Training curves for task set \#1.}
    \vspace{-6mm}
    \label{fig:traning_curves}
\end{figure}

One of the key features of DiscoSyn is the ability to simultaneously learn policies and synergies. To evaluate its performance, we compare against a sequential baseline that produces synergies via dimensionality reduction on pre-collected solutions of each task. 
For this baseline, we first generate task solutions by training independent polices for each task, and collect trajectories using the learned RL agents.
Then, we apply dimensionality reduction to these data; we use PCA as a linear method to compare against the linear version of DiscoSyn, and a neural-network-based auto-encoder as a non-linear method to compare against the non-linear version of DiscoSyn.
After applying the chosen dimensionality reduction method to extract synergies, we finally then attempt to train a policy that operates on the resulting low dimensional action space to learn each task. 
\begin{center}
    \begin{table}
            \vspace{3mm}

    \begin{tabular}{ c|c c c c } 
    
      & Valve0 & Valve1 & Valve2 & Valve3 \\ 
      \hline\\[-3.5mm]
      \hline\\[-3mm]
      PCA4 &\xmark &\xmark &\xmark &\xmark\\
      AE4 &\cmark &\xmark &\cmark & \cmark\\ 
      PCA6 &\cmark &\xmark &\xmark &\cmark\\ 
      AE6 &\cmark &\xmark &\cmark &\cmark \\
      DiscoSyn3-L & \cmark&\cmark &\cmark &\cmark \\
      DiscoSyn3-NL &\cmark &\cmark &\cmark &\cmark \\
      DiscoSyn4-L & \cmark & \cmark &\cmark &\cmark \\
      DiscoSyn4-NL &\cmark &\cmark &\cmark &\cmark \\
    %   DiscoSyn6-L &\cmark &\cmark &\cmark &\cmark \\
    %   DiscoSyn6-NL &\cmark &\cmark &\cmark &\cmark \\
    \end{tabular}
    \caption{Task set \#1 results. Each row shows the results of a method and the name of the method represent: the underlying mechanism to extract synergies, the number of synergies and the type of synergy model (L for linear and NL for non-linear). For example, DiscoSyn3-L represent using DiscoSyn with 3 linear synergies. }
    \vspace{-2mm}
    \label{tab:valve_results1}
    \end{table}
\end{center}

\vspace{-5mm}
\subsection{Task sets}
\label{sec:tasks}
We apply our method to learn manipulation tasks to 20-DoF simulated Shadow hand. We design the two task sets to be different across different characterstics of an MDP and to require the hand to interact with different objects.

\subsubsection{Task set \#1} This set contains four tasks that only vary on the transition model and have the same reward function. Specifically, we ask the hand to turn counter-clockwise different types of valves (with different number of handles). 
\subsubsection{Task set \#2} This set contains four tasks that vary on both the transition model and the reward function: dice reorienting, weight pulling, valve turning, and screwing. We design this task set to require the manipulation in different degrees of freedom. For example, the dice has 6 DoF, while the screw only has a hinge and sliding joints. 

\subsection{Performance}

As shown in Table \ref{tab:valve_results1}, DiscoSyn can learn all the tasks in task set \#1 using $3$ synergies with both linear and non-linear synergies\footnote{\label{fn:result_pointer} For more details about polices and synergy models, please find more training curves, detailed task descriptions and videos on: \href{https://roamlab.github.io/discosyn/}{https://roamlab.github.io/discosyn/}.}. On the other hand, for the more challenging task set \#2, our method learns all the tasks using $6$ synergies but fails to tackle the whole task set using $4$ linear synergies. With non-linear synergies, DiscoSyn can find a low-dimensional subspace that can allow for policy learning for all four tasks in task set \#2. 

\begin{center}
    \begin{table}
    \begin{tabular}{ c|c c c c } 
      & \head{0.6cm}{Dice} & \head{0.6cm}{Valve} & \head{1.0cm}{Weight Pulling} & Screwing \\ 
      \hline \\[-3.5mm]
      \hline \\[-3mm]
      PCA4 &\xmark &\xmark &\xmark &\cmark \\
      AE4 &\cmark &\xmark &\xmark &\cmark \\ 
      PCA6 &\cmark &\xmark &\xmark &\cmark\\  
      AE6 &\cmark &\xmark &\xmark &\cmark\\ 
      DiscoSyn4-L & \cmark&\cmark &\xmark &\xmark \\
      DiscoSyn4-NL &\cmark &\cmark &\cmark &\cmark \\
      DiscoSyn6-L &\cmark &\cmark &\cmark &\cmark \\
      DiscoSyn6-NL &\cmark &\cmark &\cmark &\cmark \\
    \end{tabular}
    \caption{Task set \#2 results.}
    \label{tab:valve_results2}
    \vspace{-6mm}
    \end{table}
\end{center}

\vspace{-6mm}
As shown in Table \ref{tab:valve_results1}, the sequential baseline with linear (PCA) synergies only learns $2$ out of $4$ tasks from task set \#1, even when using $6$ synergies.
On the other hand, the sequential baseline with non-linear (AE) synergies can perform $3$ out of $4$ tasks with $4$ synergies.
For taskset \#2, the PCA baseline only learns 2 out of 4 tasks with 6 synergies, and only learns one task with 4 synergies. 
In both task sets, the AE performance does not degrade with a decreased number of dimensions of the latent space, which can be explained by the large capacity of AE models.

% \vspace{-3mm}
\subsection{Learning unseen tasks with learned synergies}
In this experiment, we test if the learned synergies can be leveraged for learning new tasks. 
We design new test tasks for both sets: 1. turning a differently-shaped (cylindrical) valve; 2. turning a known valve in a different direction (clockwise); 3. using a screwdriver in a different direction (top-down). We use the cylindrical valve to test if the synergy model can be used to interact with novel-shape objects, the CW valve turning task to see if the synergy model can provide rich learning signals for another task with a seen object, and the top-down screwing to test if the synergies help learning task with different dynamics.

We find that, with learned synergies, an agent can learn all of these tasks efficiently (representative runs of the learned policy can be found in the video accompanying the submission). We attribute this to the learned synergies providing rich interaction between the manipulator and the objects and hence rich reward signals that lead to fast learning of unseen tasks.
\begin{figure}
    % \vspace{-2mm}
    \includegraphics[width=0.9\linewidth]{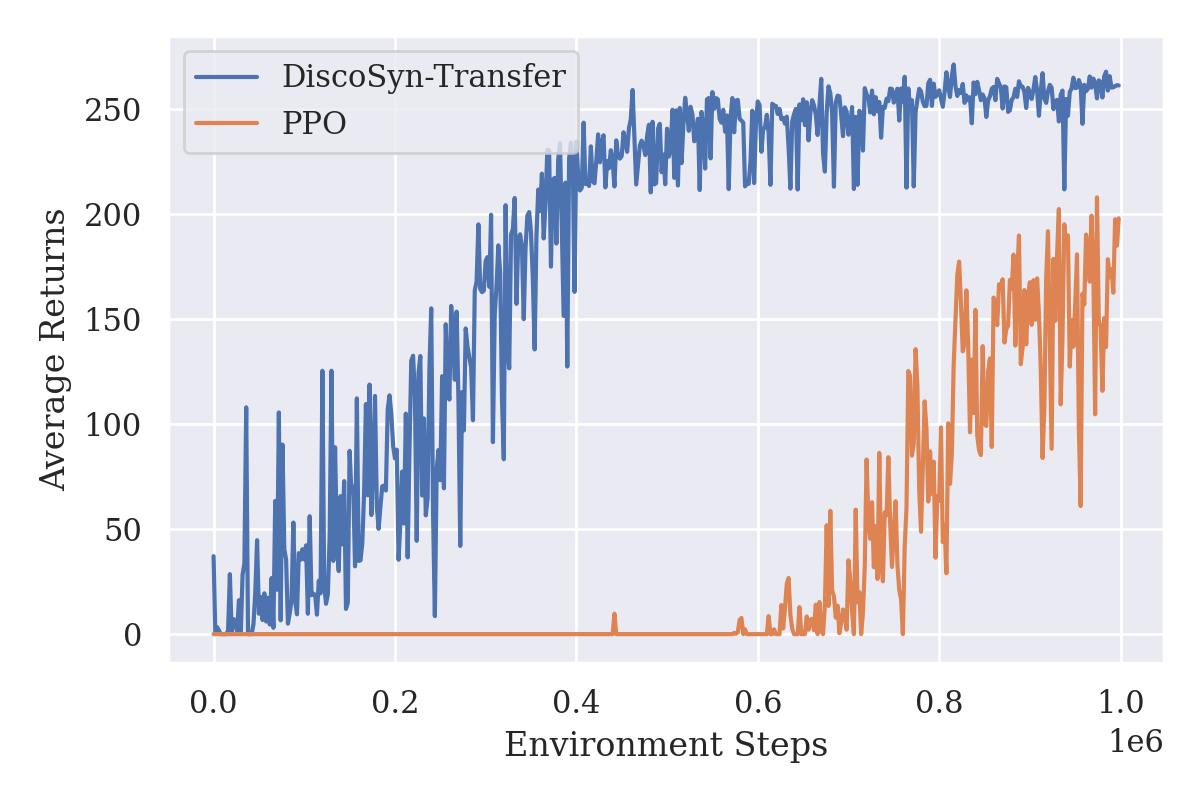}
    \vspace{-3mm}
    \caption{Training curves of learning the sparse goal-condition valve turning with a pretrained synergy model (in blue) and in full dimension action space with standard RL (in orange).}
    \label{fig:transfer}
    \vspace{-6mm}
\end{figure}

While our synergy models can be transferred to unseen tasks, we further test whether the synergies learned allows for efficient explorations. We design a sparse reward task that requires the hand to turn the valve to a specific goal joint position. The agent is only rewarded by $1$ if the distance between the valve joint position and the target is smaller than a threshold $\sigma$; otherwise the reward is $0$. As shown in Figure \ref{fig:transfer}, an agent using learned synergies observes rewards within a few time steps and starts learning the task, while a PPO agent operating on the full dimensional action space requires more than 40,000 steps to see the first reward signals.

%% file: text/results.tex
% \begin{figure}
%     \centering
%     \includegraphics[width=\linewidth]{figures/task_based_ts1.pdf}
%     \caption{}
%     \label{fig:ts1}
% \end{figure}
% \begin{figure}
%     \centering
%     \includegraphics[width=\linewidth]{figures/task_based_ts2.pdf}
%     \caption{Low dimensional actions selected by policy for task set \#2.}
%     \label{fig:ts2}
% \end{figure}

\vspace{-1mm}

\section{Discussion}
\vspace{-1mm}

\textit{1) Can DiscoSyn discover synergies while simultaneously learning tasks?} Our results show that we can extract $3$-, $4$-, and $6$-synergy spaces, and our policies learn to accomplish all tasks in the experimental sets. Furthermore, given the structure of our framework, the lower-dimensional action space always explains $100\%$ of the variance of the high-dimensional actions. 

On the other hand, applying PCA on independently learned policies only achieves explained variance of $62.8\%$, $50.8\%$ and $41.5\%$ using $6$, $4$ and $3$ respectively for task set \#1, and only slightly higher variance for task set \#2. This implies that the actions learned from standard RL do not result in a linear structure. When forced to learn a policy confined to the already-defined lower dimensional space, the sequential baseline fails to solve all tasks. 

Figure \ref{fig:handposes} presents some of the hand postures our framework learns and shows how our learned synergies are used in the training and test tasks. Our policy uses hand postures from the synergy space to interact with the object and also to transition to the next gait cycle.

% \begin{figure}[h]
%      \centering
%      \begin{subfigure}[b]{0.5\linewidth}
%          \centering
%          \includegraphics[width=\linewidth]{figures/task_based_ts1.pdf}
%          \caption{Task Set \#1}
%          \label{fig:ts1}
%      \end{subfigure}
%      \begin{subfigure}[b]{0.485\linewidth}
%          \centering
%          \includegraphics[width=\linewidth]{figures/task_based_ts2.pdf}
%          \caption{Task Set \#2}
%          \label{fig:ts2}
%     \end{subfigure}
%         \caption{Low dimensional actions selected by policy for task set \#1 (left) and task set \#2 (right).}
%     \label{fig:task-based-vis}
% \end{figure}

\begin{figure}[t]
     \centering
    \includegraphics[width=\linewidth]{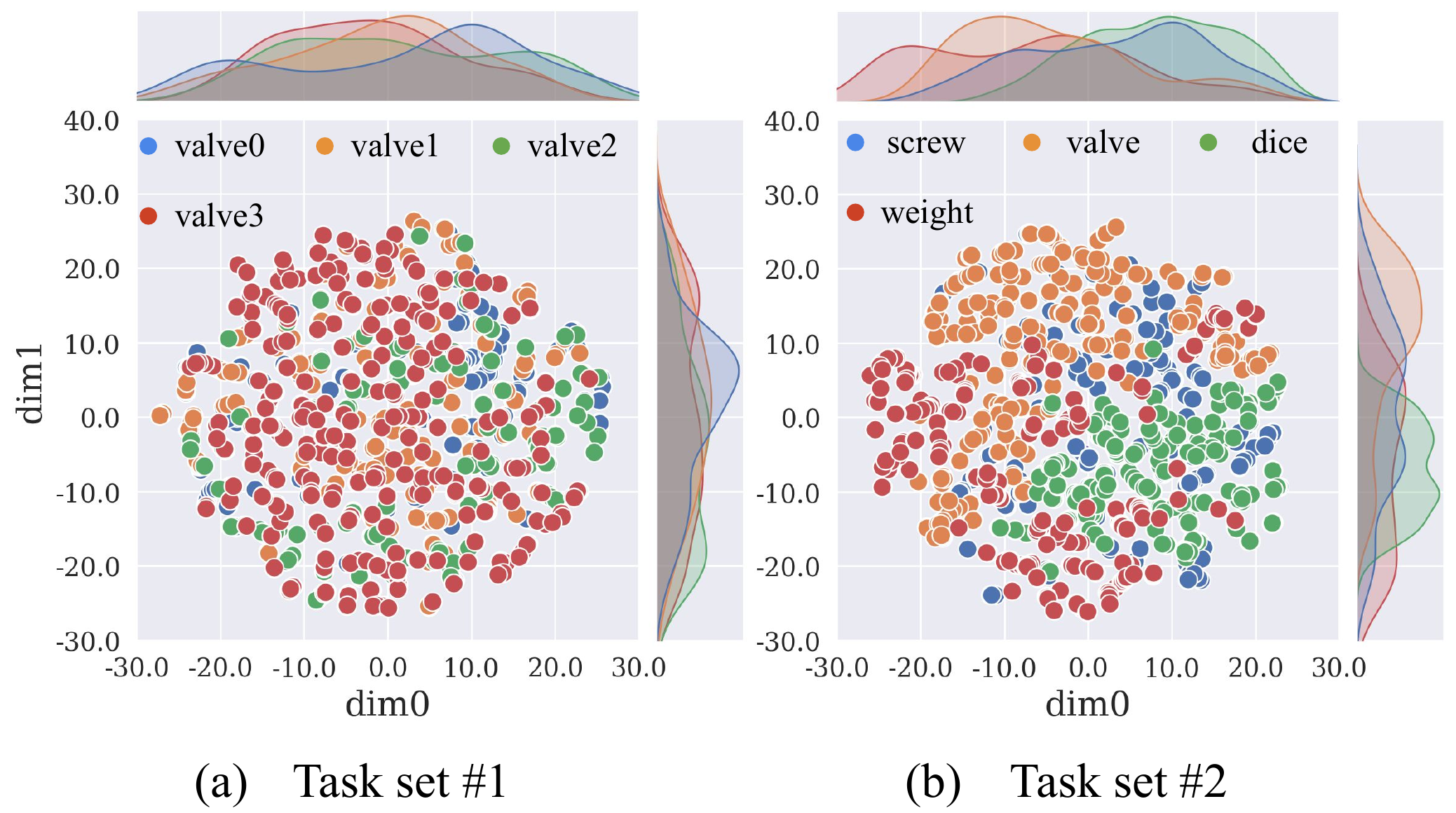}
    \label{fig:transfer_efficiency}
    \vspace{-4mm}
    \caption{t-SNE visualization of low dimensional actions selected by our policy for task set \#1 (a) and task set \#2 (b).}
    \vspace{-2mm}
\label{fig:task-based-vis}
\end{figure}

\textit{2) How is DiscoSyn affected by the training tasks?} Our results show that the diversity of a task set does influence the learning of the synergy model. This is expected: if a task set contains many different tasks, the capacity of $z$ space needs to be larger. Our experimental results show that we cannot learn a linear synergy model with $4$ synergies for task set \#2 while DiscoSyn can use as few as 3 synergies for task set \#1.
We also visualize how each task utilizes the $z$ space for both task sets. Since the tasks in task set \#1 are similar, they share a large portion of the $z$ space. On the other hand, in task set \#2, which contains more diverse tasks, the task-dependent policies tend to occupy different parts of the $z$ space. For example, the valve turning and dice re-orientation only overlap in a small space.

% \subsection{Can we use learned synergy to learn unseen task efficiently?}

% To test if learned synergies allow for efficient explorations, we design a sparse reward task: we ask the robotic hand to do goal-conditioned valve turning. 

% \begin{figure*}
%     \centering
%     \includegraphics[width=\linewidth]{figures/synergy_evo.pdf}
%     \caption{Training curves for taskset \#1. We also visualize the min and max hand pose of each synergy dimension for several iterations. \todo{Add one for PCA and AE in the margin below the curves probably? Probably only visualize disco3L and disco3NL. Remove the box and only show x and y axis}}
%     \label{fig:traning_curves}
% \end{figure*}

%% file: text/conclusion.tex
\vspace{-2mm}
\section{conclusions}

Our results show that DiscoSyn is able to learn synergies that are effective for multiple manipulation tasks, while simultaneously learning policies that operate on this low dimensional action space in an end-to-end manner. We also demonstrate that the learned synergy model can be reused for unseen task and allows for efficient exploration during transfer. Compared to a classic pipeline that extracts synergies on existing control data, our method can use fewer synergies to solve multiple tasks. We note that, in its current stage, DiscoSyn is limited by requiring a pre-defined number of latent dimensions. In the future, we hope to equip DiscoSyn with the ability to discover the dimensionality of the smallest synergy space needed for a set of task, and also to apply its results to the design of underactuated robotic hands.

%% file: references.bib
@article{nester2019hansynergydata,
author = {Jarque-Bou, Nestor and Scano, Alessandro and Atzori, Manfredo and Müller, Henning},
year = {2019},
month = {05},
pages = {},
title = {Kinematic synergies of hand grasps: a comprehensive study on a large publicly available dataset},
volume = {16},
journal = {Journal of NeuroEngineering and Rehabilitation},
doi = {10.1186/s12984-019-0536-6}
}

@article{marco1998toolusepose,
author = {Santello, Marco and Flanders, Martha and Soechting, John},
year = {1998},
month = {12},
pages = {10105-15},
title = {Postural Hand Synergies for Tool Use},
volume = {18},
journal = {The Journal of neuroscience : the official journal of the Society for Neuroscience},
doi = {10.1523/JNEUROSCI.18-23-10105.1998}
}

@article{mason2002reachgraspsyn,
author = {Mason, C.R. and Gomez, J.E. and Ebner, T.J.},
year = {2002},
month = {01},
pages = {2896-910},
title = {Hand Synergies During Reach-to-Grasp},
volume = {86},
journal = {Journal of neurophysiology},
doi = {10.1152/jn.2001.86.6.2896}
}

@article{Liu2016AnalysisOH,
  title={Analysis of Hand and Wrist Postural Synergies in Tolerance Grasping of Various Objects},
  author={Y. Liu and Li Jiang and Dapeng Yang and Hong Liu},
  journal={PLoS ONE},
  year={2016},
  volume={11}
}

@article{Patel2017HandGS,
  title={Hand Grasping Synergies As Biometrics},
  author={Vrajeshri Patel and P. Thukral and Martin K. Burns and I. Florescu and R. Chandramouli and R. Vinjamuri},
  journal={Frontiers in Bioengineering and Biotechnology},
  year={2017},
  volume={5}
}

@INPROCEEDINGS{todorov2004handsynergies,
    author={Todorov, E. and Ghahramani, Z.}, 
    booktitle={The 26th Annual International Conference of the IEEE Engineering in Medicine and Biology Society},
    title={Analysis of the synergies underlying complex hand manipulation},
    year={2004},
    volume={2},
    number={},
    pages={4637-4640},
    doi={10.1109/IEMBS.2004.1404285}
}

@article{ciocarlie2009posture,
    author = {Ciocarlie, Matei and Allen, Peter},
    year = {2009},
    month = {06},
    pages = {851-867},
    title = {Hand Posture Subspaces for Dexterous Robotic Grasping},
    volume = {28},
    journal = {I. J. Robotic Res.},
    doi = {10.1177/0278364909105606}
}

@inproceedings{suarez2015dualarm,
author = {Suarez, Raul and Rosell, Jan and Garcia, Nestor},
year = {2015},
month = {06},
pages = {5655-5661},
title = {Using synergies in dual-arm manipulation tasks},
volume = {2015},
journal = {Proceedings - IEEE International Conference on Robotics and Automation},
doi = {10.1109/ICRA.2015.7139991}
}

@INPROCEEDINGS{luck2017groups,
  author={Luck, Kevin Sebastian and Ben Amor, Heni},
  booktitle={2017 IEEE/RSJ International Conference on Intelligent Robots and Systems (IROS)}, 
  title={Extracting bimanual synergies with reinforcement learning}, 
  year={2017},
  volume={},
  number={},
  pages={4805-4812},
  doi={10.1109/IROS.2017.8206356}}

@article{openai2018dex,
  author    = {OpenAI and
               Marcin Andrychowicz and
               Bowen Baker and
               Maciek Chociej and
               Rafal J{\'{o}}zefowicz and
               Bob McGrew and
               Jakub W. Pachocki and
               Jakub Pachocki and
               Arthur Petron and
               Matthias Plappert and
               Glenn Powell and
               Alex Ray and
               Jonas Schneider and
               Szymon Sidor and
               Josh Tobin and
               Peter Welinder and
               Lilian Weng and
               Wojciech Zaremba},
  title     = {Learning Dexterous In-Hand Manipulation},
  journal   = {CoRR},
  volume    = {abs/1808.00177},
  year      = {2018},
  url       = {http://arxiv.org/abs/1808.00177},
  archivePrefix = {arXiv},
  eprint    = {1808.00177},
  bibsource = {dblp computer science bibliography, https://dblp.org}
}

@article{nagabandi2019pddm,
  author    = {Anusha Nagabandi and
               Kurt Konolige and
               Sergey Levine and
               Vikash Kumar},
  title     = {Deep Dynamics Models for Learning Dexterous Manipulation},
  journal   = {CoRR},
  volume    = {abs/1909.11652},
  year      = {2019},
  url       = {http://arxiv.org/abs/1909.11652},
  archivePrefix = {arXiv},
  eprint    = {1909.11652},
  bibsource = {dblp computer science bibliography, https://dblp.org}
}

@article{kroemer2019manipulation,
  author    = {Oliver Kroemer and
               Scott Niekum and
               George Dimitri Konidaris},
  title     = {A Review of Robot Learning for Manipulation: Challenges, Representations,
               and Algorithms},
  journal   = {CoRR},
  volume    = {abs/1907.03146},
  year      = {2019},
  url       = {http://arxiv.org/abs/1907.03146},
  archivePrefix = {arXiv},
  eprint    = {1907.03146},
  bibsource = {dblp computer science bibliography, https://dblp.org}
}

@inproceedings{
hausman2018learning,
title={Learning an Embedding Space for Transferable Robot Skills},
author={Karol Hausman and Jost Tobias Springenberg and Ziyu Wang and Nicolas Heess and Martin Riedmiller},
booktitle={International Conference on Learning Representations},
year={2018},
url={https://openreview.net/forum?id=rk07ZXZRb},
}

@inproceedings{yu2019meta,
  title={Meta-World: A Benchmark and Evaluation for Multi-Task and Meta Reinforcement Learning},
  author={Tianhe Yu and Deirdre Quillen and Zhanpeng He and Ryan Julian and Karol Hausman and Chelsea Finn and Sergey Levine},
  booktitle={Conference on Robot Learning (CoRL)},
  year={2019},
  eprint={1910.10897},
  archivePrefix={arXiv},
  primaryClass={cs.LG},
  url={https://arxiv.org/abs/1910.10897}
}

@article{sodhani2021contextrl,
  author    = {Shagun Sodhani and
               Amy Zhang and
               Joelle Pineau},
  title     = {Multi-Task Reinforcement Learning with Context-based Representations},
  journal   = {CoRR},
  volume    = {abs/2102.06177},
  year      = {2021},
  url       = {https://arxiv.org/abs/2102.06177},
  archivePrefix = {arXiv},
  eprint    = {2102.06177},
  bibsource = {dblp computer science bibliography, https://dblp.org}
}

@INPROCEEDINGS{Tanaka2003mtmdps,
  author={Tanaka, F. and Yamamura, M.},
  booktitle={Proceedings 2003 IEEE International Symposium on Computational Intelligence in Robotics and Automation. Computational Intelligence in Robotics and Automation for the New Millennium (Cat. No.03EX694)}, 
  title={Multitask reinforcement learning on the distribution of MDPs}, 
  year={2003},
  volume={3},
  number={},
  pages={1108-1113 vol.3},
  doi={10.1109/CIRA.2003.1222152}}

@inproceedings{Calandriello2014sparsemtrl,
 author = {Calandriello, Daniele and Lazaric, Alessandro and Restelli, Marcello},
 booktitle = {Advances in Neural Information Processing Systems},
 editor = {Z. Ghahramani and M. Welling and C. Cortes and N. Lawrence and K. Q. Weinberger},
 pages = {},
 publisher = {Curran Associates, Inc.},
 title = {Sparse Multi-Task Reinforcement Learning},
 url = {https://proceedings.neurips.cc/paper/2014/file/94c7bb58efc3b337800875b5d382a072-Paper.pdf},
 volume = {27},
 year = {2014}
}

@article{borsa2016mtrl,
  author    = {Diana Borsa and
               Thore Graepel and
               John Shawe{-}Taylor},
  title     = {Learning Shared Representations in Multi-task Reinforcement Learning},
  journal   = {CoRR},
  volume    = {abs/1603.02041},
  year      = {2016},
  url       = {http://arxiv.org/abs/1603.02041},
  eprinttype = {arXiv},
  eprint    = {1603.02041},
  bibsource = {dblp computer science bibliography, https://dblp.org}
}

@inproceedings{DEramo2020Sharing,
title={Sharing Knowledge in Multi-Task Deep Reinforcement Learning},
author={Carlo D'Eramo and Davide Tateo and Andrea Bonarini and Marcello Restelli and Jan Peters},
booktitle={International Conference on Learning Representations},
year={2020},
url={https://openreview.net/forum?id=rkgpv2VFvr}
}

@article{peter2008rlmotorskills,
author = {Peters, Jan and Schaal, Stefan},
year = {2008},
month = {06},
pages = {682-},
title = {Reinforcement learning of motor skills with policy gradients},
volume = {21},
journal = {Neural Networks},
doi = {10.1016/j.neunet.2008.02.003}
}

@article{theodorou2010reinforcementklo,
  title={Reinforcement learning of motor skills in high dimensions: A path integral approach},
  author={E. Theodorou and J. Buchli and S. Schaal},
  journal={2010 IEEE International Conference on Robotics and Automation},
  year={2010},
  pages={2397-2403}
}

@article{Daniel2016reps,
  author  = {Christian Daniel and Gerhard Neumann and Oliver Kroemer and Jan Peters},
  title   = {Hierarchical Relative Entropy Policy Search},
  journal = {Journal of Machine Learning Research},
  year    = {2016},
  volume  = {17},
  number  = {93},
  pages   = {1-50},
  url     = {http://jmlr.org/papers/v17/15-188.html}
}

@article{schulman2017ppo,
  author    = {John Schulman and
               Filip Wolski and
               Prafulla Dhariwal and
               Alec Radford and
               Oleg Klimov},
  title     = {Proximal Policy Optimization Algorithms},
  journal   = {CoRR},
  volume    = {abs/1707.06347},
  year      = {2017},
  url       = {http://arxiv.org/abs/1707.06347},
  archivePrefix = {arXiv},
  eprint    = {1707.06347},
  bibsource = {dblp computer science bibliography, https://dblp.org}
}

@article{Napier1956ThePM,
  title={The prehensile movements of the human hand.},
  author={J. Napier},
  journal={The Journal of bone and joint surgery. British volume},
  year={1956},
  volume={38-B 4},
  pages={
          902-13
        }
}

@article{iberall1997humanprehension,
author = {Thea Iberall},
title ={Human Prehension and Dexterous Robot Hands},
journal = {The International Journal of Robotics Research},
volume = {16},
number = {3},
pages = {285-299},
year = {1997},
doi = {10.1177/027836499701600302},

URL = { 
        https://doi.org/10.1177/027836499701600302
    
},
eprint = { 
        https://doi.org/10.1177/027836499701600302
    
}
}
